\documentclass[10pt,twocolumn,letterpaper]{article}

\usepackage{cvpr}              % To produce the CAMERA-READY version
\usepackage{multirow}
\usepackage{makecell}
\usepackage{stfloats}
\definecolor{cvprblue}{rgb}{0.21,0.49,0.74}
\usepackage[pagebackref,breaklinks,colorlinks,allcolors=cvprblue]{hyperref}
\usepackage{cuted} % 导言区
\def\paperID{15448} % *** Enter the Paper ID here
\def\confName{CVPR}
\def\confYear{2026}

\title{ OmniFood8K: Single-Image Nutrition Estimation via Hierarchical Frequency-Aligned Fusion }

\author{
Dongjian Yu$^{1}$ \quad
Weiqing Min$^{2,3}$ \quad
Qian Jiang$^{1}$ \quad
Xing Lin$^{1}$ \quad
Xin Jin$^{1} $\thanks{Corresponding author} \quad
Shuqiang Jiang$^{2,3}$\\[0.5em]
$^{1}$School of Software, Yunnan University, China    
$^{2}$State Key Laboratory of AI Safety,   \\ Institute of Computing Technology,
 Chinese Academy of Sciences, China \\
$^{3}$University of Chinese Academy of Sciences, China
 }

\begin{document}

\maketitle
% \footnotetext[1]{* Corresponding author.}

\begin{abstract}
Accurate estimation of food nutrition plays a vital role in promoting healthy dietary habits and personalized diet management.
Most existing food datasets primarily focus on Western cuisines and lack sufficient coverage of Chinese dishes, which restricts accurate nutritional estimation for Chinese meals.
Moreover, many state-of-the-art nutrition prediction methods rely on depth sensors, restricting their applicability in daily scenarios.
To address these limitations, 
we introduce OmniFood8K, a comprehensive multimodal dataset comprising 8,036 food samples, each with detailed nutritional annotations and multi-view images.
In addition, to enhance models’ capability in nutritional prediction, we construct NutritionSynth-115K, a large-scale synthetic dataset that introduces compositional variations while preserving precise nutritional labels.
Moreover, we propose an end-to-end framework for nutritional prediction from a single RGB image. 
First, we predict a depth map from a single RGB image and design the Scale-Shift Residual Adapter (SSRA) to refine it for global scale consistency and local structural preservation.
Second, we propose the Frequency-Aligned Fusion Module (FAFM) to hierarchically align and fuse RGB and depth features in the frequency domain. Finally, we design a Mask-based Prediction Head (MPH) to emphasize key ingredient regions via dynamic channel selection for more accurate prediction.
Extensive experiments on multiple datasets demonstrate the superiority of our method over existing approaches.
Project homepage: \href{https://yudongjian.github.io/OmniFood8K-food/}{https://yudongjian.github.io/OmniFood8K-food/}

\end{abstract}    

\section{Introduction}
\label{sec:intro}

\begin{figure}[t]  
    \centering
    \includegraphics[width=0.42\textwidth]{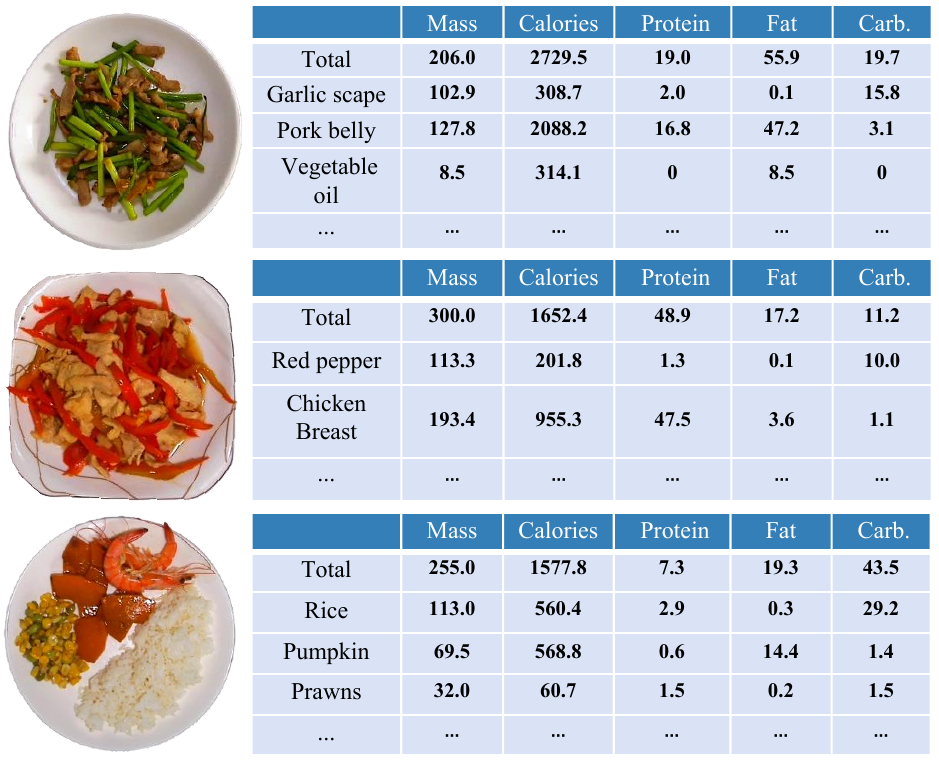}
    \caption{ Representative examples from the OmniFood8K dataset.} % 图注
    \label{image-1}  % 引用标签
\end{figure}

With the improvement of living standards and the diversification of dietary habits, nutritional assessment of food has become a key research focus in both public health and personal health management \cite{min2019survey, wang2021smart, quan2025machine}. Accurate estimation of food energy and nutrient intake plays a vital role in preventing chronic diseases such as obesity, diabetes, and cardiovascular disorders \cite{konstantakopoulos2023review, shen2020machine}. However, traditional nutritional assessment methods \cite{whitton2024accuracy, o2024dietary, singh2025minimum} often rely on manual recording or expert estimation, which are not only time-consuming and labor-intensive but also prone to subjective bias \cite{jakstas2026psychometric, mozaffarian2026food}. 
This makes accurate and consistent evaluation challenging, particularly in real-world daily dietary applications \cite{feng2024ingredient, qi2025advancing}.

Recent advances in computer vision and multimodal learning have significantly advanced the development of automated food nutrition assessment  \cite{foods11213429, SHAO2023136309, shao2023end, zheng2024artificial, deng2026joint, deng2026ihmambasr}.
Deep learning techniques have shown great potential in automatically recognizing, segmenting, analyzing food images to estimate food mass, volume, and nutritional composition \cite{wang2022review, foods12234293, shonkoff2023ai}. Moreover, leveraging multimodal features such as textual descriptions and sensor data can further enhance estimation accuracy and robustness, facilitating personalized nutrition monitoring and healthcare systems \cite{thames2021nutrition5k, chinthala2026food}.

% 修改如下内容
Despite these advances, current methods still encounter two fundamental limitations that constrain their practical usability.
\textbf{(1) Data Limitation.} Existing datasets are heavily biased toward Western cuisines, with limited representation of Chinese food. The inherent diversity, complex ingredient composition, and non-standardized preparation of Chinese food make reliable annotation and quantitative nutritional analysis particularly challenging.
\textbf{(2) Algorithmic Limitation.} 
% Many advanced approaches depend on depth information for accurate nutrition estimation. 
Many advanced approaches rely on depth maps captured by depth cameras for accurate nutrition estimation.
Nevertheless, in most real-world contexts, food photographs are taken with conventional RGB cameras rather than depth sensors, thereby restricting the practical adoption of these methods.

% These limitations underscore the need for robust and practical food nutrition assessment methods that can effectively handle both Western and Chinese foods.

% To address two limitations, this paper makes two key contributions:
% First, we construct a comprehensive, multimodal food nutrition dataset designed for real-world applications. This dataset spans the entire food preparation process, covering raw ingredient images, ingredient mass information, structured recipe descriptions, complete cooking video records, and multi-view photos of the finished dishes (including stereo images), all paired with detailed nutritional labels. Unlike previous fragmented or “isolated” datasets, our dataset is unified and process-oriented, with strong causal links across each stage—from ingredient preparation and cooking steps to the final presentation. This design not only reflects realistic dietary scenarios but also provides a unified foundation for multiple research tasks such as nutrition estimation, image segmentation, 3D reconstruction, and recipe generation, thereby laying the groundwork for building practical and user-friendly intelligent nutrition analysis methods (see Table 1).

\begin{figure*}
    \centering
       \includegraphics[width=0.95\textwidth]{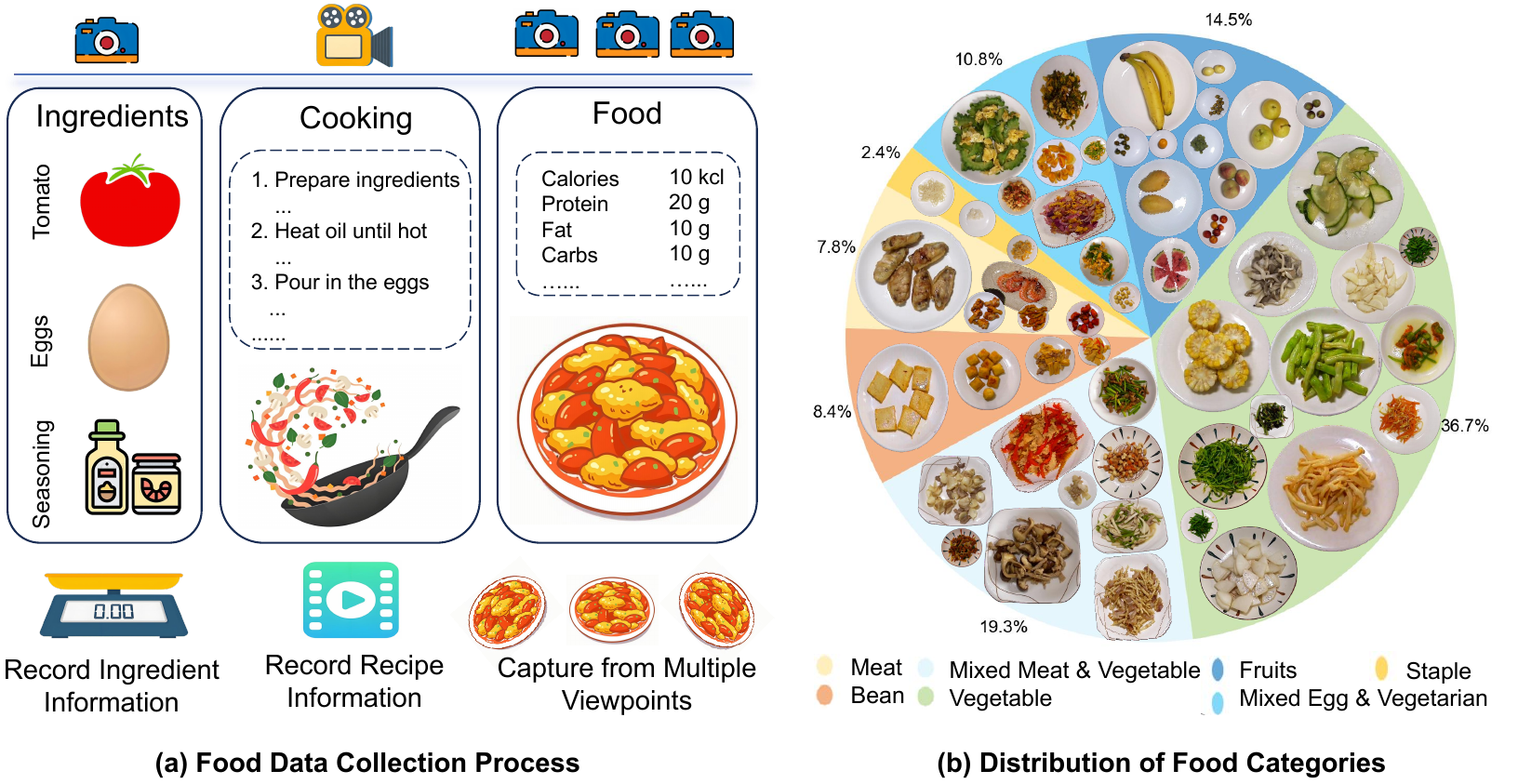} % 图片路径
    \caption{Overview of the OmniFood8K dataset: data collection process and category distribution.} % 图注
    \label{image-2}  % 引用标签
\end{figure*}

To address these two limitations, this work makes two key contributions.
First, we introduce OmniFood8K, a comprehensive multimodal food nutrition dataset designed for real-world scenarios, as shown in Figure \ref{image-1}. 
The dataset covers the entire food preparation process, including raw ingredient images with mass information, structured recipe descriptions, full cooking videos, and multi-view images of the finished dishes, all paired with detailed nutritional annotations, as illustrated in Figure \ref{image-2}.
Unlike existing datasets, OmniFood8K employs a unified, process-oriented design that preserves causal links from ingredient preparation to final presentation and provides comprehensive nutritional annotations.
In addition, to further improve the model’s nutritional prediction capability, we construct NutritionSynth-115K, a large-scale synthetic dataset with compositional diversity and precise nutritional annotations.

Second, we propose an end-to-end framework that predicts nutritional information directly from a single RGB image. Compared with depth-sensor-based approaches, our method is more generalizable and scalable for real-world dietary assessment.
Specifically, we first employ a pre-trained depth estimation model \cite{yang2024depth} to predict the depth information of food. To correct scale bias and local structural errors, we design a Scale-Shift Residual Adapter (SSRA) for consistent global and local calibration.
We then employ a Frequency-Aligned Fusion Module (FAFM) to hierarchically fuse RGB and adapted depth features, aligning multi-modal representations in the frequency domain for enhanced cross-modal learning.
Finally, a Mask-based Prediction Head (MPH) leverages dynamic channel selection and region-aware attention to emphasize key ingredient regions, improving nutrition prediction accuracy.

Overall, our contributions are summarized as follows:
\begin{itemize}

\item 
% We introduce OmniFood8K, a comprehensive multimodal food dataset with detailed nutritional annotations. It encompasses ingredient information, recipe descriptions, cooking videos, and multi-view images of final food, thereby offering a comprehensive foundation for research on multimodal nutrition estimation.
We introduce OmniFood8K, a comprehensive multimodal food dataset with detailed nutritional annotations, covering ingredients, recipes, cooking videos, and multi-view food images.
Additionally, we construct NutritionSynth-115K, a large-scale synthetic dataset featuring diverse food compositions and precise nutritional annotations.
\item We propose an end-to-end framework for predicting food nutritional information from a single RGB image. 
It incorporates a Scale-Shift Residual Adapter (SSRA) to refine depth estimation and achieve consistent calibration of both global scale and local geometry.

\item We further design a Frequency-Aligned Fusion Module (FAFM) to hierarchically fuse RGB and depth features in the frequency domain, and a Mask-based Prediction Head (MPH) that dynamically selects informative channels, thereby enhancing the accuracy of nutritional prediction.

\item Extensive experiments on multiple datasets demonstrate that our method outperforms state-of-the-art approaches, validating its effectiveness and scalability for food nutrition assessment.
\end{itemize}

\section{Related work}
\label{sec:formatting}

%-------------------------------------------------------------------------
\subsection{Food Datasets}

Recent advances in food computing have led to the release of numerous publicly available datasets \cite{gui2024navigating, arrighi2025explainable}, which have substantially promoted research across multiple domains, such as food recognition, cross-modal analysis, and nutritional estimation \cite{esmaeily2024step, Romero-Tapiador_2025_CVPR, bianco20252d, tanabe2025reasoning}.
Representative datasets such as Food-101 \cite{bossard2014food}, VIREO Food-172 \cite{chen2016deep}, ISIA Food-500 \cite{min2020isia}, and Food2K \cite{min2023large} cover diverse food categories and have been widely adopted for recognition and classification tasks \cite{zhang2023deep, zhang2025recent, 10743395}.
Additionally, Recipe1M \cite{salvador2017learning} aligns food images with textual recipes, serving as a fundamental resource for cross-modal retrieval and recipe generation tasks.
Nutrition5k \cite{thames2021nutrition5k} includes detailed nutritional annotations and has been instrumental in advancing research on food nutrition assessment.
However, its primary focus on Western dishes, with limited coverage of complex cuisines such as Chinese food, constrains its general applicability.
FoodSeg103 \cite{wu2021large} is a dataset designed for food image segmentation, aiming to facilitate fine-grained understanding of food components.
MetaFood3D \cite{chen2024metafood3d} is a multimodal dataset comprising 637 3D food objects across 108 categories, accompanied by detailed nutritional information and diverse visual modalities.
The FastFood dataset \cite{qi2025advancing} collects food images and their corresponding nutritional information from official fast-food brand websites, providing valuable data for nutrition analysis and related studies.

% ChinaMartFood-109 is a nutrition estimation dataset covering 109 commonly available Chinese market foods, including grains, vegetables, meats, and fruits. However, since daily meals typically consist of mixed ingredients, the applicability of this dataset for real-world nutrition prediction is limited.

\subsection{Food Nutrition Assessment}
Traditional nutrition assessment methods are time-consuming, labor-intensive, and often limited in predictive accuracy \cite{chotwanvirat2024advancements, varzakas2024global, robinson2026concerns}. With advances in computer vision, AI-based approaches for nutrition prediction have increasingly emerged  \cite{foods12234293, jiao2024rode, foods11213429, thames2021nutrition5k, wang2023coarse, shi2021rice}.
NR \textit{et al.} \cite{nr2022framework} employed CNN to automatically recognize food and predict its nutritional information.
Swin-Nutrition \cite{foods11213429} introduced an efficient, non-destructive AI-based framework for accurately predicting multiple food nutrient components using Swin Transformer features.
Vinod \textit{et al.} \cite{9874714} integrated energy density maps with depth information to effectively improve calorie estimation accuracy.
Shao \textit{et al.} \cite{shao2023end} proposed estimating food energy through 3D shape reconstruction from a single-view image, achieving promising results.
Feng \textit{et al.} \cite{feng2024ingredient} proposed a method that incorporates ingredient information to complement visual features, thereby improving the accuracy of nutrition prediction and achieving promising results.
RoDE \cite{jiao2024rode} introduced a multi-expert framework to improve the accuracy and efficiency of large-scale food multi-modal models for tasks such as nutrition estimation.
Ma \cite{ma2025food} \textit{et al.} proposed FBFPN, a network that fuses RGB and depth images via a bidirectional feature pyramid to improve nutrition estimation accuracy.

%-------------------------------------------------------------------------

\section{OmniFood8K Dataset Construction}

\subsection{Motivation}
% 为什么要做这个数据集
Existing nutrition datasets primarily focus on Western foods and lack sufficient representation of complex cuisines such as Chinese dishes, limiting their applicability to accurate nutritional estimation across diverse dietary contexts.
To address these limitations, we introduce OmniFood8K, a comprehensive multimodal dataset for practical nutrition assessment, supporting diverse tasks such as nutrition estimation, food image generation, ingredient recognition, and recipe generation.

\subsection{Data Collection}
% 数据从哪来，怎么采集
The dataset comprises 8,036 food samples, representing a wide range of commonly consumed Chinese dishes.
Data were collected from real-world cooking scenarios, including raw ingredients, structured recipes, cooking videos, and final dish presentations.
The data acquisition process, illustrated in Figure~\ref{image-2}, follows a structured and systematic workflow. 
Raw ingredients were first weighed and photographed, followed by video recording of the cooking process and textual documentation of the recipes.
Finally, the prepared food was captured simultaneously from six viewpoints at a horizontal height of 50 cm \cite{https://doi.org/10.1155/2014/489757}.
This height approximates the typical elbow level of an adult, making the setup convenient and practical for everyday use in real-life scenarios.
Two cameras were positioned directly above the food, while the remaining four were placed at the front, back, left, and right sides.

\begin{table*}[ht]
    % \large
    \renewcommand{\arraystretch}{1.1}
    \centering
    \small
    \caption{Comparison between the proposed dataset and existing food datasets.}
    \label{table_dataset}
     \begin{tabular}{lccccccccc}
    \hline
    \makecell{Dataset} & \makecell{Categories} & \makecell{Size} & \makecell{Ingredient \\ Image} & \makecell{Food \\ Image} & \makecell{Multi-view \\ Food Image} & \makecell{Recipe} & \makecell{Cooking \\ Video} & \makecell{Nutritional \\ Info} \\
    \hline
    Food101   \cite{bossard2014food}   & 101  & 101,000  & - & $\checkmark$ & - & - & - & - \\
    VIREO Food-172 \cite{chen2016deep}      & 172  & 110,241 & - & $\checkmark$ & - & $\checkmark$ & - & - \\
    Recipe1M \cite{salvador2017learning}     & 1480    & 1,047,000   & -  &$\checkmark$ & - & $\checkmark$ & - & - \\
    Yummly-66K \cite{8059846}  & 10    & 66,615   & - &  $\checkmark$ &- & $\checkmark$ & - & - \\
    ISIA Food-200  \cite{min2019ingredient} & 200  & 197,323 & -  & $\checkmark$ &- & - & - & - \\
    ISIA Food-500 \cite{min2020isia}    & 500  & 399,726  & -  & $\checkmark$ & - &- & - & - \\
    Food2K    \cite{min2023large}    & 2000 & 1,036,564      & -  & $\checkmark$ & - &- & - & - \\
    YouCook2  \cite{zhou2018towards}   & 89   & 2000        & $\checkmark$ &$\checkmark$ & - & $\checkmark$ & $\checkmark$  & - \\
    Menu-Match  \cite{7045971} & 41    & 646     & - & $\checkmark$ &- & - & - & $\checkmark$ \\
    Fang \textit{et al.}  \cite{fang2015single} & 3    & 45     & - & $\checkmark$ &- & - & - & $\checkmark$ \\
    Nutrition5k  \cite{thames2021nutrition5k} & -    & 5066    & - & $\checkmark$ &$\checkmark$ & - & - & $\checkmark$ \\
    MetaFood3D \cite{chen2024metafood3d} & 108    & 637     & - & $\checkmark$ &$\checkmark$ & - & - & $\checkmark$ \\
     OmniFood8K (\textbf{Ours})         & 165  & 8,036    & $\checkmark$ & $\checkmark$ & $\checkmark$ & $\checkmark$ & $\checkmark$ & $\checkmark$ \\
    
    NutritionSynth-115K (\textbf{Ours})         & 165  & 115,000    & $\checkmark$ & $\checkmark$ & - & $\checkmark$ & $\checkmark$ & $\checkmark$ \\
    \hline
    \end{tabular}
\end{table*}

\subsection{Annotation}
% 标签怎么做
For each food scene, ingredients were first photographed and their corresponding weights recorded to enable precise computation of nutritional information. 
% Based on these measurements, detailed nutritional annotations were generated for each scene, including calorie, protein, fat, and carbohydrate contents, which were subsequently validated using standardized nutritional databases. 
Based on the recorded ingredient weights, the nutritional contents of each scene, including calories, protein, fat, and carbohydrates, were calculated by referencing official nutritional databases to ensure accurate estimation of each food item’s nutrient composition.
Recipes were documented using structured textual descriptions, while the entire cooking process was comprehensively recorded on video. Additionally, key steps in each recipe were captured as corresponding images to support multimodal analysis.
% The final food were captured simultaneously from six different viewpoints to obtain multi-view images.

\subsection{Dataset Characteristics}
% 数据集特点和优势
% OmniFood8K provides nutritional annotations for 8,036 food scenes, accompanied by diverse multimodal data, including ingredient images with weight information, structured recipe descriptions, cooking process videos, multi-view images of the final food, and food nutritional data, as shown in Table~\ref{tab:data_overview}.
OmniFood8K provides nutritional annotations for 8,036 food samples, accompanied by diverse multimodal data, whose types and corresponding applications are summarized in Table~\ref{tab:data_overview}.
\begin{table}[t]
  \caption{Data types and corresponding applications in the OmniFood8K dataset.}
  \label{tab:data_overview}
  \centering
  \small
  \setlength{\tabcolsep}{4pt}
  \renewcommand{\arraystretch}{1.25}
  \begin{tabular}{@{}p{2.2cm} p{6cm}@{}}
    \toprule
    \textbf{Data Type} & \textbf{Applications} \\
    \midrule
    Ingredient Images  & Ingredient Recognition, Detection, Quantity Estimation. \\ \hline
    Recipe Descriptions & Recipe Generation, Cross-Modal Retrieval, Cooking Reasoning. \\ \hline
    Cooking Process Videos & Action Recognition, Cooking Stage Understanding, Multimodal Alignment with Recipes. \\ \hline
    Multi-view Images of Food & Food Recognition, Detection, Image Synthesis, 3D-Aware Food Understanding. \\ \hline
    Food Nutritional Data & Nutrition Prediction, Dietary Assessment, Health-oriented Analysis. \\
    \bottomrule
  \end{tabular}
\end{table}
By spanning the entire food preparation pipeline, OmniFood8K serves as a comprehensive resource for research in food computing and nutritional analysis.
Table \ref{table_dataset} presents a comparison of OmniFood8K with existing food datasets across multiple characteristics.
Notably, OmniFood8K is the first dataset to cover the entire food preparation process, from raw ingredients to finished food. Its main features are summarized as follows:
\begin{itemize}
    \item \textbf{Multimodal}: OmniFood8K includes images, structured recipe descriptions, cooking videos, and detailed nutritional annotations.
    % \item \textbf{Full pipeline coverage}: OmniFood8K covers the entire food preparation process, from raw ingredients with images and weights, through recipe descriptions and cooking videos, to final food captured from multiple viewpoints.

    \item \textbf{Full Pipeline Coverage}: OmniFood8K spans the complete cooking process, from annotated ingredient images and weights to multi-view final food images, accompanied by recipe descriptions and cooking videos.

    \item \textbf{Causal Relationship}: OmniFood8K preserves the causal relationship from raw ingredients to finished dishes through recipe descriptions and cooking videos.

     \item \textbf{Accuracy:} Unlike other datasets \cite{jiao2024rode, qi2025advancing} that rely on web-scraped labels or LLM-generated annotations, OmniFood8K provides labels based on real-world weight measurements and standardized nutritional database calculations, ensuring higher accuracy.

\end{itemize}

\begin{figure*}
    \centering
    \includegraphics[width=\textwidth]{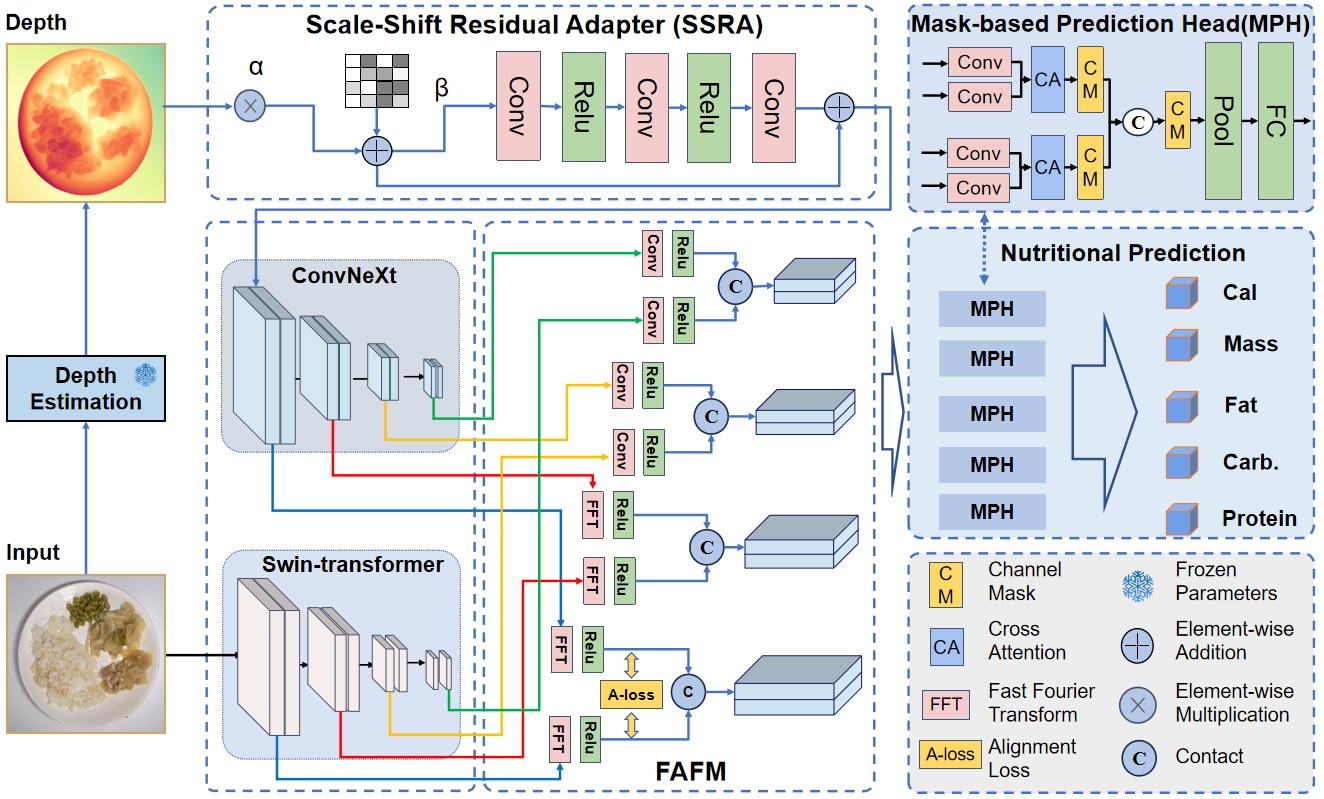} % 图片路径
    \caption{Overview of the proposed method.
The figure illustrates the overall pipeline of our method, consisting of three proposed modules: the Scale-Shift Residual Adapter (SSRA), Frequency-Aligned Fusion Module (FAFM), and Mask-based Prediction Head (MPH).} % 图注
    \label{network_pipeline}  % 引用标签
\end{figure*}

\subsection{NutritionSynth-115K}
We performed foreground segmentation on OmniFood8K food images to isolate individual items. These segmented items were subsequently cropped or recombined according to their original proportions, allowing the synthesized samples to faithfully reflect the nutritional composition of each meal. Through this process, we constructed a large-scale and controllable synthetic dataset, named NutritionSynth-115K, which contains 115,000 images with precise nutritional annotations.

\section{Method}

% The overall architecture of our method is shown in Figure \ref{network_pipeline}.
% A single RGB image is first processed by a depth prediction model to generate a depth map, which is refined by SSRA for global and local correction. The RGB image and the refined depth map are then fed into Swin Transformer and ConvNeXt to extract multi-scale features, which are further fused via PMMR. Finally, the MPH module predicts the nutritional information, enabling end-to-end nutrition estimation.
The overall architecture of the proposed method is illustrated in Figure~\ref{network_pipeline}.
A single RGB image is first fed into a depth estimation model to produce an initial depth map, which is subsequently refined using the proposed Scale-Shift Residual Adapter (SSRA) module for both global and local corrections.
The RGB image and the refined depth map are then used to extract multi-scale features, which are fused by the Frequency-Aligned Fusion Module (FAFM) through frequency-domain alignment of multimodal representations. Finally, the Mask-based Prediction Head (MPH) predicts the nutritional information, forming an end-to-end framework for nutrition estimation.

\subsection{Scale-Shift Residual Adapter (SSRA)}
To ensure practical applicability in daily scenarios, we focus on nutrition estimation from a single RGB image.
Since RGB images inherently lack explicit depth cues, we employ a monocular depth estimation model \cite{yang2024depth} to infer depth information.
To further mitigate estimation errors, we design a lightweight Scale-Shift Residual Adapter (SSRA) that efficiently refines both global and local structures.

Specifically, at the global level, we introduce a learnable scaling factor $\alpha$ and shift parameter $\beta$ to align the overall distribution of the predicted depth map $d_{mono}$:
\begin{equation}
d_{global} = \alpha \cdot d_{mono} + \beta .
\end{equation}
This step effectively mitigates the common issues of scale ambiguity and offset errors in monocular depth prediction. 
Building upon this, a shallow convolutional residual refinement module $f_{\theta}(\cdot)$ is employed to enhance local structural details and correct geometric distortions. 
The refinement process can be expressed as:
\begin{equation}
d_{res} = f_\theta(d_{global}) .
\end{equation}
The final adapted depth is then obtained as:
\begin{equation}
d_{out} = d_{global} + d_{res} .
\end{equation}
By combining global alignment with local refinement, the proposed SSRA ensures consistency in overall depth estimation while improving local precision, thus enhancing the accuracy of  nutrition estimation tasks.

\subsection{Frequency-Aligned Fusion Module (FAFM)}
% To effectively leverage the complementary information from RGB and depth modalities and preserve information integrity during multi-modal feature fusion, we design a \textbf{Frequency-Aligned Fusion Module (FAFM)}. This module performs multi-frequency decomposition and fusion to enhance cross-modal feature representation.
To effectively integrate complementary RGB and depth information while preserving feature consistency, we propose a Frequency-Aligned Fusion Module (FAFM). This module fuses multi-frequency features and enforces cross-modal consistency via an inter-modal alignment loss.

Given RGB and depth feature maps $r, d \in \mathbb{R}^{C \times H \times W}$, the module first applies a 2D Fast Fourier Transform (FFT) to obtain the frequency representations $R_f = \mathcal{F}(r)$ and $D_f = \mathcal{F}(d)$. Based on a predefined frequency threshold $\kappa$, a low-frequency mask $M_L$ is constructed to separate low-frequency (global structure) and high-frequency (local detail) components:
\begin{equation}
R_L = \mathcal{F}^{-1}(R_f \odot M_L), \quad R_H = \mathcal{F}^{-1}(R_f \odot (1 - M_L)),
\end{equation}
\begin{equation}
D_L = \mathcal{F}^{-1}(D_f \odot M_L), \quad D_H = \mathcal{F}^{-1}(D_f \odot (1 - M_L)).
\end{equation}
The high-frequency and low-frequency components from both modalities are then fused via element-wise addition:
\begin{equation}
F_H = R_H + D_H, \quad F_L = R_L + D_L .
\end{equation}
The fused high-frequency and low-frequency features are concatenated along the channel dimension and passed through a learnable convolutional layer to adaptively compress and integrate them, producing the final cross-modal feature representation:
\begin{equation}
O = \text{Conv}([F_H, F_L]) .
\end{equation}
In this design, high-frequency components enhance texture and local details, while low-frequency components preserve global structure, achieving Multi-Frequency Fusion.

To enhance semantic consistency between RGB and depth features, we introduce an inter-modal alignment loss $\mathcal{L}_{align}$, defined as:
\begin{equation}
\label{eq:align}
\mathcal{L}_{align} = - \frac{1}{N} \sum_{i=1}^{N} \log \frac{\exp(\text{sim}(\mathbf{f}^r_i, \mathbf{f}^d_i) / \tau)}{\sum_{j=1}^{N} \exp(\text{sim}(\mathbf{f}^r_i, \mathbf{f}^d_j) / \tau)},
\end{equation}
where $\mathbf{f}^r_i$ and $\mathbf{f}^d_i$ are the global feature vectors of the $i$-th sample in RGB and depth modalities, $\text{sim}(\cdot,\cdot)$ denotes cosine similarity, $N$ is the batch size, and $\tau$ is a temperature parameter. 
This loss encourages features from the same sample across modalities to be close, while separating features from different samples, thus improving cross-modal representation for subsequent fusion.

% \subsection{Progressive Multi-Modal Refinement (PMMR)}

% To effectively exploit heterogeneous cues from different modalities, we propose a Progressive Multi-Modal Refinement module (PMMR). Unlike existing approaches that mostly focus on dual- or tri-modal fusion, our design explicitly incorporates RGB, depth, CLIP, and DINO features into a unified hierarchical framework.

% Specifically, the lower layers (Layer 1–2) primarily integrate RGB and depth features, which preserve fine-grained spatial and geometric details. The higher layers (Layer 3–4) further incorporate CLIP embeddings (via 1×1 projection) and DINO tokens (via 1D-to-2D projection and interpolation), thus injecting cross-modal semantic priors and self-supervised global knowledge into the representation. In this way, the model progressively refines its fused features from low-level structural cues to high-level semantic abstractions.

% To ensure compatibility with the UNet decoder while avoiding feature redundancy, we introduce a dimension-controlled fusion strategy, where concatenated features from multiple modalities are projected and channel-trimmed to fixed dimensions (192, 384, 768, 1536). This makes our framework both expressive and computationally efficient.

% Overall, our PMMR provides a progressive and modality-aware fusion mechanism, bridging visual appearance, geometric depth, cross-modal semantics, and global self-supervised knowledge, which significantly enhances downstream performance in food nutrition estimation tasks.

\subsection{Mask-based Prediction Head (MPH)}

To fully leverage multi-scale features and emphasize key ingredient regions, we propose the Mask-based Prediction Head (MPH). This module takes feature maps from different sources as input, including multi-scale RGB features and semantic features. First, MPH applies convolutional operations and adaptive average pooling to each feature map to extract high-dimensional representations and unify the spatial dimensions to a fixed size, ensuring the controllability and stability of subsequent fusion operations. Next, MPH employs a Cross-Attention mechanism to enable information interaction between RGB and semantic features, allowing the model to capture correlations across different modalities. This is followed by Gated Fusion, which further reweights the fused features to highlight the most critical channels. To further enhance the focus on key ingredient regions, MPH introduces a dynamic channel mask (Channel Mask), which automatically selects the most informative channels, thereby preserving the features most relevant to the target nutritional information during fusion. Finally, the RGB and semantic features are integrated through a global fusion module and processed by global pooling and a fully connected layer to produce the predicted nutritional values.

% \subsection{Loss Function}

% In our nutrient prediction task, we directly adopt the Dynamic Task-weighted Loss proposed in \cite{10.1145/3746027.3755750}, which dynamically adjusts task weights according to the difficulty of each nutrient prediction. The loss is defined as:

% \begin{equation}
% \text{Loss} = \sum_{i=1}^{n} w_i \cdot \text{PMAE}(\text{nutri}[i]) ,
% \end{equation}
% where $w_i$ denotes the dynamic weight of the $i$-th task, computed from the current learning difficulty using KPI and updated with smoothing:

% \begin{equation}
% \begin{split}
% \text{KPI}_i &= \frac{1}{1 - \text{PMAE}(\text{nutri}[i])}, \\
% w_i^{(t)} &= \alpha \cdot \text{KPI}_i^{(t)} + (1-\alpha) \cdot w_i^{(t-1)} .
% \end{split}
% \end{equation}
% Here, $n$ is the number of nutrient components, and PMAE is the Percentage Mean Absolute Error.

\subsection{Loss Function}

In the nutrient prediction task, we adopt a Dynamic Task-weighted Loss \cite{10.1145/3746027.3755750} for nutrient estimation, which dynamically adjusts task weights according to the current prediction difficulty. The nutrient prediction loss is formulated as:

\begin{equation}
\mathcal{L}_{\text{nutri}} = \sum_{i=1}^{n} w_i \cdot \text{PMAE}(\text{nutri}[i]),
\end{equation}
where $n$ is the number of nutrient components, $w_i$ is the dynamically updated weight for the $i$-th nutrient task, and PMAE denotes the Percentage Mean Absolute Error. Specifically, each task $i$ corresponds to a nutrient component such as Calorie, Mass, Fat, Carbohydrate, or Protein. The dynamic weights are updated according to the task difficulty via a KPI metric:

\begin{equation}
\begin{split}
\text{KPI}_i &= \frac{1}{1 - \text{PMAE}(\text{nutri}[i])}, \\
w_i^{(t)} &= \alpha \cdot \text{KPI}_i^{(t)} + (1-\alpha) \cdot w_i^{(t-1)},
\end{split}
\end{equation}
where $t$ denotes the training iteration, and $\alpha$ is a smoothing factor controlling the update momentum, which is empirically set to 0.3 in our experiments.

To enhance cross-modal representation consistency, we further introduce an inter-modal alignment loss $\mathcal{L}_{\text{align}}$ (Eq.~\ref{eq:align}), applied to fused RGB–Depth features.
Finally, the total loss for training the nutrient prediction model is defined as:
\begin{equation}
\mathcal{L}_{\text{total}} = \mathcal{L}_{\text{nutri}} + \lambda \, \mathcal{L}_{\text{align}},
\end{equation}
where $\lambda$ balances the contribution of the inter-modal alignment loss. This formulation enables joint optimization of nutrient prediction and cross-modal feature alignment, thereby improving the model's predictive performance in multimodal fusion.

\section{Experiments}
\subsection{Experimental Setup}

\begin{table*}[]
    \centering
    \caption{Comparison of PMAE (\%) values of different baselines on the OmniFood8K dataset.}
    \label{t_2}
    \small
    \renewcommand\arraystretch{1.15}
    \setlength{\tabcolsep}{5.6mm}
    \begin{tabular}{@{}llcccccc@{}}
        \toprule
        \textbf{Input} & \textbf{Method} & \textbf{Calories} & \textbf{Mass} & \textbf{Fat} & \textbf{Carb.} & \textbf{Protein} & \textbf{Mean} \\ 
        \midrule
        \multirow{10}{*}{RGB images} 
        & Inception V3 \cite{szegedy2016rethinking} & 17.2 & 10.0 & 17.8 & 13.8 & 20.5 & 16.0 \\
        & ResNet-50 \cite{he2016deep} & 10.2 & 5.1 & 11.0 & 11.1 & 14.8 & 10.4 \\
        & ResNet-101 \cite{he2016deep} & 10.9 & 6.0 & 12.2 & 9.1 & 15.2 & 10.6 \\
        & DenseNet-121 \cite{Huang_2017_CVPR} & 12.1 & 7.4 & 13.7 & 10.5 & 18.2 & 12.4 \\
        & DenseNet-161 \cite{Huang_2017_CVPR} & 12.2 & 7.6 & 12.7 & 12.6 & 16.3 & 12.3 \\
        & EfficientNet \cite{tan2019efficientnet} & 27.1 & 10.4 & 28.3 & 38.9 & 40.0 & 28.9 \\
        & MobileNetV3 \cite{howard2019searching} & 17.1 & 8.5 & 18.5 & 21.2 & 27.1 & 18.5 \\
        & Swin-Transformer \cite{liu2021swin} & 27.7 & 15.1 & 29.6 & 42.6 & 41.0 & 31.2 \\
        & ConvNeXt \cite{liu2022convnet} & 10.1 & 5.6 & 10.8 & 9.5 & 13.3 & 9.8 \\
        & ConvNeXt V2 \cite{Woo_2023_CVPR} & 15.5 & 6.4 & 11.3 & 9.3 & \underline{\textbf{12.6}} & 11.2 \\
        % \midrule
        & \textbf{Ours} & \underline{\textbf{8.8}} & \underline{\textbf{4.5}} & \underline{\textbf{9.5}} & \underline{\textbf{8.0}} & \underline{\textbf{12.6}} & \underline{\textbf{8.6}} \\
        \bottomrule
    \end{tabular}
\end{table*}

\begin{table*}[t]
\centering
\caption{Comparison of PMAE (\%) values of different baselines on the Nutrition5k dataset.}
\label{t_3}
\small
\renewcommand\arraystretch{1.15}
\setlength{\tabcolsep}{3.5mm}
\begin{tabular*}{\textwidth}{@{\extracolsep{\fill}}llcccccc@{}}
\toprule
\textbf{Input} & \textbf{Method} & \textbf{Calories} & \textbf{Mass} & \textbf{Fat} & \textbf{Carb.} & \textbf{Protein} & \textbf{Mean} \\
\midrule

\multirow{6}{*}{RGB images} 
 & Google-Nutrition-rgb \cite{thames2021nutrition5k} & 26.1 & 18.8 & 34.2 & 31.9 & 29.5 & 29.1 \\
 & Coarse-to-Fine Nutrition \cite{wang2023coarse} & 24.1 & 19.4 & 36.0 & 32.1 & 33.5 & 29.0 \\
 & Swin-Nutrition \cite{foods11213429} & 16.2 & 13.7 & 24.9 & 21.8 & 25.4 & 20.4 \\
 & Portion-Nutrition \cite{shao2023end} & 15.8 & - & - & - & - & - \\
 & RoDE \cite{jiao2024rode} & 52.4 & 38.4 & 67.1 & 47.8 & 53.9 & 51.9 \\
 & DPF-Nutrition \cite{foods12234293} & 14.7 & 10.6 & 22.6 & 20.7 & 20.2 & 17.8 \\
\midrule

\multirow{13}{*}{RGB-D images}
 & CMX \cite{10231003} & 21.8 & 20.7 & 34.8 & 37.0 & 33.2 & 29.5 \\
 & HINet \cite{BI2023109194} & 24.5 & 25.2 & 43.4 & 39.9 & 38.8 & 34.3 \\
 & CDINet \cite{zhang2021cross} & 21.1 & 20.4 & 37.1 & 37.1 & 32.8 & 29.7 \\
 & DEFNet \cite{zhou2022defnet} & 32.7 & 34.2 & 48.9 & 40.3 & 43.8 & 39.9 \\
 & TriTransNet \cite{liu2021tritransnet} & 22.1 & 20.1 & 37.5 & 34.8 & 38.0 & 30.5 \\
 & Deliver \cite{zhang2023delivering} & 29.5 & 25.9 & 48.3 & 47.7 & 46.1 & 39.5 \\
 & Google-Nutrition-rgbd \cite{thames2021nutrition5k} & 18.8 & 18.9 & \underline{\textbf{18.1}} & 23.8 & 20.9 & 20.1 \\
 & Domain Adaptation-Nutrition \cite{9874714} & 16.8 & - & - & - & - & - \\
 & RGB-D Net \cite{SHAO2023136309} & 15.0 & 10.8 & 23.5 & 22.4 & 21.0 & 18.5 \\
 & IMIR-Net \cite{nian2024ingredient} & 14.5 & 10.4 & 21.8 & 20.4 & 20.0 & 17.4 \\
 & FBFPN \cite{ma2025food} & \underline{\textbf{14.0}} & 10.3 & 22.6 & 19.5 & 20.2 & 17.3 \\
\midrule

\multirow{1}{*}{RGB images} 
 & \textbf{Ours} & 14.1 & \underline{\textbf{10.2}} & 21.0 & \underline{\textbf{18.9}} & \underline{\textbf{18.4}} & \underline{\textbf{16.5}} \\
\bottomrule
\end{tabular*}
\end{table*}

\textbf{1) Evaluation Metrics.}
The  percentage of mean absolute error (PMAE) is adopted to evaluate the prediction performance of our method. This metric measures the relative deviation between predicted and true values, with smaller values indicating more accurate predictions.
PMAE is defined as:
\begin{align}
\text{MAE} &= \frac{1}{N} \sum_{i=1}^{N} |y_t - y_p| , \\
\text{PMAE} &= \frac{ \text{MAE} }{\frac{1}{N} \sum_{i=1}^{N} y_t} \times 100\% ,
\end{align}
where \( y_t \) denotes the true value, \( y_p \) denotes the predicted value, and \( N \) denotes the total number of samples.
A smaller value indicates higher prediction accuracy and better model performance.
\\
\textbf{2) Baseline Methods.}
To evaluate the effectiveness of our proposed method, we conduct a comparison with recent state-of-the-art approaches. 
% On the OmniFood8K dataset, we select representative models including Inception V3  \cite{szegedy2016rethinking}, ResNet \cite{he2016deep}, DenseNet \cite{Huang_2017_CVPR}, EfficientNet \cite{tan2019efficientnet}, MobileNetV3 \cite{howard2019searching}, Swin-Transformer \cite{liu2021swin}. 
On the OmniFood8K dataset, we select representative models including   \cite{szegedy2016rethinking, he2016deep, Huang_2017_CVPR, tan2019efficientnet, howard2019searching, liu2021swin}. 
For the Nutrition5k dataset, we compared our method with recent approaches, including Google-Nutrition \cite{thames2021nutrition5k}, Portion-Nutrition \cite{shao2023end}, Coarse-to-Fine Nutrition \cite{wang2023coarse}, Swin-Nutrition \cite{foods11213429}, DPF-Nutrition \cite{foods12234293}, RGB-D Net \cite{SHAO2023136309}, Domain Adaptation-Nutrition \cite{9874714}, IMIR-Net \cite{nian2024ingredient}, FBFPN \cite{ma2025food}. We also compare our method with prior image fusion approaches \cite{BI2023109194, zhang2021cross, 10231003, zhang2023delivering, zhou2022defnet, liu2021tritransnet} .\\
\textbf{3) Implementation Details.}
All models were implemented in PyTorch and trained on an NVIDIA A100 GPU. The Adam optimizer was used with a weight decay of 1e-5 to prevent overfitting. A cosine annealing scheduler adjusted the learning rate during training. Models were trained for 150 epochs with a batch size of 8.
The OmniFood8K and Nutrition5k datasets are split into training and testing sets at ratios of 7:3 and 5:1 \cite{SHAO2023136309, foods12234293}, respectively.

\begin{table*}[t]
\centering
\caption{Ablation study of the proposed modules on the Nutrition5k dataset.}
\label{t_ab1}
\small
\renewcommand\arraystretch{1.15}
\setlength{\tabcolsep}{4.2mm} % 调整列间距让两端对齐更自然
\begin{tabular*}{\textwidth}{@{\extracolsep{\fill}}lccccccccc@{}}
\toprule
\textbf{Baseline} & \textbf{FAFM} & \textbf{SSRA} & \textbf{MPH} & \textbf{Calories} & \textbf{Mass} & \textbf{Fat} & \textbf{Carb.} & \textbf{Protein} & \textbf{Mean} \\
\midrule
\checkmark &  &  &  & 14.8 & 11.3 & 23.6 & 20.1 & 19.7 & 17.9 \\
\checkmark & \checkmark &  &  & 14.4 & \textbf{10.2} & 21.7 & 19.7 & 19.2 & 17.0 \\
\checkmark & \checkmark & \checkmark &  & 14.2 & 10.4 & 21.9 & 19.2 & 18.6 & 16.8 \\
\checkmark & \checkmark & \checkmark & \checkmark & \textbf{14.1} & \textbf{10.2} & \textbf{21.0} & \textbf{18.9} & \textbf{18.4} & \textbf{16.5} \\
\bottomrule
\end{tabular*}
\end{table*}

\subsection{ Experimental Results}
To validate the effectiveness of the proposed method, we conduct experiments on multiple food nutrition estimation datasets. Table \ref{t_2} presents the comparison with different baseline methods on our proposed OmniFood8K dataset. It can be observed that our method achieves the best performance across multiple metrics, with the lowest PMAE, indicating superior prediction accuracy.
Table \ref{t_3} reports the comparison results on the Nutrition5k dataset. Our method outperforms existing state-of-the-art approaches on the PMAE metric across multiple nutrient components, achieving the best overall performance. 
Notably, even when compared with a number of state-of-the-art methods that directly leverage RGB-D inputs, our method still achieves superior performance, further demonstrating its effectiveness and robustness.

As shown in Tables \ref{t_2} and \ref{t_3}, our method achieves a lower PMAE on OmniFood8K than on Nutrition5k, demonstrating superior predictive accuracy.
This improvement can be attributed to the larger scale of OmniFood8K, which contains 8,036 images (5,600 used for training) compared to Nutrition5k’s 3,500 images (2,800 for training) \cite{feng2024ingredient, SHAO2023136309}. The roughly twofold increase in training samples exposes the model to more diverse food appearances and nutritional compositions, enhancing feature learning and prediction performance.

\subsection{Ablation Study}
To evaluate the contribution of each component in our method, we conduct an ablation study on the Nutrition5k dataset. We gradually add the proposed components and observe the resulting performance changes. As shown in Table \ref{t_ab1}, the model’s performance steadily improves with the inclusion of each component, demonstrating the effectiveness of our method.

\subsection{Pretraining on NutritionSynth-115K}
To enhance the performance of our method on the OmniFood8K dataset, it is first pretrained on NutritionSynth-115K. This pretraining enables the model to capture rich food features and multimodal representations, providing a robust initialization for subsequent fine-tuning on OmniFood8K. As shown in Table \ref{table_pretraining}, pretraining significantly improves predictive performance, demonstrating its effectiveness in enhancing generalization.
\begin{table}[]
    % \large
    \small
    \begin{center}
        \caption{Evaluation of our method pre-trained on NutritionSynth-115K and fine-tuned on OmniFood8K.}
        \label{table_pretraining}
        % 行间距  列间距  设置如下两行
        \renewcommand\arraystretch{1.2} 
        \setlength{\tabcolsep}{1.4mm}  % 这里去掉多余的 `{`
        \scalebox{1}{
            \begin{tabular}{ccccccccccc}
                \hline
                Method   & Calories & Mass & Fat & Carb. & Protein & Mean \\  
                \hline
                Ours & 8.8 
                 & \textbf{4.5} &  9.5& 8.0 & 12.6 &  8.6 \\  \hline 
  
                + Pre-trained &\textbf{ 8.0} & \textbf{4.5 }&\textbf{9.0} &\textbf{7.4} & \textbf{10.0} & \textbf{7.8} \\   
                
                \hline
            \end{tabular}
        }
    \end{center}
\end{table}

\section{Conclusion}
% To enable scalable and user-friendly nutritional estimation, we propose an end-to-end framework that predicts nutritional information from a single RGB image. We construct OmniFood8K, comprising 8,036 real-world food scenes with multi-view images, ingredient weights, recipes, and nutritional annotations. In addition, we introduce NutritionSynth-115K, a large-scale synthetic dataset that augments real food images with compositional variations while retaining accurate nutritional labels. 
% We introduce the Scale-Shift Residual Adapter (SSRA) to calibrate monocular depth prediction globally and locally, ensuring scale consistency and structural accuracy. The Frequency-Aligned Fusion Module (FAFM) hierarchically fuses RGB and depth features in the frequency domain, with lower layers emphasizing spatial cues and higher layers integrating cross-modal semantics. Finally, the Mask-based Prediction Head (MPH) dynamically selects informative channels, improving prediction accuracy. Extensive experiments across multiple datasets demonstrate that our approach consistently outperforms existing methods in nutritional prediction.
To enable practical and accessible nutritional estimation, we presented an end-to-end framework capable of predicting nutritional information directly from a single RGB image.
In addition, we introduced OmniFood8K, a comprehensive dataset containing 8,036 real-world food samples with multi-view images, ingredient weights, recipes, cooking videos, and detailed nutritional annotations.
To further improve nutritional prediction, we also introduced NutritionSynth-115K, a large-scale synthetic dataset that preserves accurate nutritional labels.
Our framework first predicts monocular depth, which is globally and locally refined by the Scale-Shift Residual Adapter (SSRA) to ensure scale consistency and preserve structural accuracy. 
The Frequency-Aligned Fusion Module (FAFM) hierarchically fuses RGB and depth features in the frequency domain, capturing both spatial and cross-modal semantic information. 
Finally, the Mask-based Prediction Head (MPH) dynamically selects informative channels to improve prediction accuracy.
Extensive experiments across multiple datasets demonstrate that our approach consistently outperforms existing methods in nutritional prediction, validating its effectiveness and generalizability.
% In future work, we plan to investigate broader multimodal cues and real-world deployment for comprehensive nutritional assessment.

% To enable scalable and user-friendly nutritional estimation, we propose an end-to-end framework that predicts nutritional information from a single RGB image. We construct OmniFood8K, comprising 8,036 real-world food scenes with multi-view images, ingredient weights, recipes, and nutritional annotations, and NutritionSynth-115K, a large-scale synthetic dataset featuring diverse food compositions with accurate nutritional labels.
% The framework integrates a Scale-Shift Residual Adapter (SSRA) for global and local depth calibration, a Frequency-Aligned Fusion Module (FAFM) for hierarchical fusion of RGB and depth features in the frequency domain, and a Mask-based Prediction Head (MPH) for dynamic channel selection, collectively enhancing prediction accuracy.
% Extensive experiments on multiple datasets demonstrate that our approach consistently outperforms existing methods in nutritional prediction.

\section*{Acknowledgments}

% This work was supported by XXXX (Grant No. XXXX). The authors would like to thank XXXX for valuable discussions and technical support.

This work was supported by the National Natural Science Foundation of China (Nos. 62261060, 62472411, 62125207), Beijing Natural Science Foundation (No. JQ24021), Yunnan Fundamental Research Projects (Nos. 202503AG380006, 202301AW070007), Yunnan Province Expert Workstations (No. 202305AF150078), Yunnan Province Special Project (No. 202403AP140021),  and Xingdian Talent Project in Yunnan Province of China.

% 闵老师 基金
% National Natural Science Foundation of China (Nos. 62472411, and 62125207), 
% Beijing Natural Science Foundation (No. JQ24021)

{
    \small
    \bibliographystyle{ieeenat_fullname}
    \bibliography{main.bib}
}


% {
%     \small
%     \bibliographystyle{ieeenat_fullname}
%     \bibliography{main}

\end{document}